\documentclass{article}
\usepackage{graphicx}
\usepackage{caption}
\usepackage{subcaption}
\usepackage{tikz}

\usepackage[nonatbib,final]{neurips_2023}
\usepackage{amsmath}
\usepackage[style=numeric,sorting=none]{biblatex}
\usepackage[utf8]{inputenc} 
\usepackage[T1]{fontenc}    
\usepackage{hyperref}       
\usepackage{url}            
\usepackage{booktabs}       
\usepackage{amsfonts}       
\usepackage{nicefrac}       
\usepackage{microtype}      
\usepackage{xcolor}         

\title{A Saliency-based Clustering Framework for Identifying Aberrant Predictions}

\author{
  Aina Tersol Montserrat\\
  SignalPET\\
  \texttt{aina.tersol@gmail.com} \\
  \And
  Alexander R. Loftus \\
  San Diego, CA \\
  \texttt{alexloftus2004@gmail.com}
    \And
  Yael Daihes \\
  SignalPET \\
  \texttt{daihesyael@gmail.com}
}

\begin{document}
\maketitle

\begin{abstract}
In machine learning, classification tasks serve as the cornerstone of a wide range of real-world applications. Reliable, trustworthy classification is particularly intricate in biomedical settings, where the ground truth is often inherently uncertain and relies on high degrees of human expertise for labeling. Traditional metrics such as precision and recall, while valuable, are insufficient for capturing the nuances of these ambiguous scenarios. Here we introduce the concept of \textit{aberrant predictions}, emphasizing that the nature of classification errors is as critical as their frequency. We propose a novel, efficient training methodology aimed at both reducing the misclassification rate and discerning aberrant predictions. Our framework demonstrates a substantial improvement in model performance, achieving a 20\% increase in precision. We apply this methodology to the less-explored domain of veterinary radiology, where the stakes are high but have not been as extensively studied compared to human medicine. By focusing on the identification and mitigation of aberrant predictions, we enhance the utility and trustworthiness of machine learning classifiers in high-stakes, real-world scenarios, including new applications in the veterinary world.
\end{abstract}

\section{Introduction}
The challenge of building robust and reliable machine learning models in the classification setting has largely gravitated towards improving accuracy, precision, and recall (\cite{Louis2017PrecisionRecallVA}, \cite{NiculescuMizil2005PredictingGP}, \cite{DeDiego2022GeneralPS}). These metrics serve as the yardstick for evaluating the performance and reliability of models in a myriad of applications. However, particularly in settings where boundaries between labels are determined based on value judgments rather than stark delineations, we begin to care about how these models made their predictions. This is particularly important in biomedical applications where label ambiguity reflects inherent uncertainty in medical diagnoses and prognoses \cite{Alpert2004}.In these scenarios, both true and false predictions could be considered valuable. This is because each prediction might capture a different facet of the underlying medical uncertainty, reflecting the ambiguity intrinsic to the labels themselves. However, while some predictions can be justified as capturing this uncertainty, others might be completely off the mark—either due to model limitations or misinterpretation of the data. In such cases, these "unjustifiable predictions" could mislead clinicians and compromise patient care. 

This paper focuses on what happens when these types of predictions are made. We define an \textit{aberrant prediction}, which refers to a significantly illogical or unjustifiable prediction made by a classifier. For instance, a study by Lechner et al. \cite{Lechner2020NeuralCP} discusses a real-world case where a self-driving car focuses on the bush on the side of the road to make its predictions. While the model may be technically `correct,' in its final action-prediction, its reasoning can confound human observers, thereby undermining the system's reliability and eroding trust in the model's decision-making process.

\subsection{Problem Statement}
    Imagine a state-of-the-art machine learning model designed to assist veterinary medical practitioners by suggesting whether fractures are present in animal X-rays. This model is well-calibrated \cite{wang2023}, ensuring that its output probabilities accurately reflect real-world likelihoods. It also performs consistently across various subgroups of data such as age, gender, and breed of the animals. Consider a specific scenario where the patient's X-ray shows an unusual pattern caused by a foreign object. In this case, the model might predict the presence of a fracture, even assigning it a high probability score, indicating strong confidence in this diagnosis. However, the model did not make this prediction based on clinical fracture features in the bone region. This would be an aberrant prediction. Note that even if the X-Ray actually has a fracture in it, the prediction is still aberrant if it is based on the wrong input features.

\subsubsection*{Definition}

In the classification setting, we have input variable $X$, a categorical variable $Y \in \{1, 2, \dots, k\}$, and a trained neural network $f$ which maps some $x$ to a categorical distribution $p = \{p_1, \dots, p_k\}$ over $k$ classes $\{y_1, \dots, y_k\}$: $f : D \rightarrow \Delta$ where $\Delta$ is the $k-1$ dimensional standard probability simplex, $D$ is the data distribution, and $\Delta = \{p \in [0,1]^k | \sum_{i=1}^k p_i = 1\}$.

A model uses discriminative features $C_x$ to make its prediction $f(x) = p$, and each class label is associated with true features $C_y$. $C_y$ is unknown to the network and inherent to the properties of the object being labeled. In an ideal network $C_x$ = $C_y$, meaning the predictions are based on the true, unknown features associated with the class label $y$.  A prediction is \textit{aberrant} if $C_x$ significantly differs from $C_y$ under some dissimilarity measure $d$:

\begin{align}
d(C_y, C_x) > \lambda, \quad \text{where } \lambda \text{is some dissimilarity threshold.}
\end{align}

Implementation-wise, the discriminative features of a classifier can be obtained from the saliency maps learned by the network \cite{Zhou2015LearningDF, Ramprasaath2019, Guo2017LIMELI, trumbelj2014ExplainingPM}. The true features can be estimated from bounding-box ground truths around the relevant regions of the image.

\subsection{Causes of Aberrant Predictions}
Aberrant predictions have a number of causes. They may arise from limitations in the training dataset, which is generally a sparse sampling of the true distribution. They could also stem from the model's architecture, which might misgeneralize based on the features it has learned. Additional sources of error could include mislabeling or the presence of correlated but non-causal features in the training data (for instance, the presence of a right-side-marker (R) every time an x-ray discovers a fracture).

From a probabilistic standpoint, the model's aberrant predictions can be understood as a limitation in capturing the full support of the data distribution, particularly long-tail events. It's crucial to note that real-world scenarios are filled with sub-optimal acquisition techniques and edge cases that may not be represented in training sets, and so most real-world applications involve a long tail distribution of edge cases.

In light of the above, the main contributions of this paper are:
\begin{enumerate}
\item{\textbf{Definition of aberrant predictions}} We propose a definition for aberrant predictions which provides a grounded underpinning for future work. This aids in distinguishing aberrant from regular predictions.
\item{\textbf{Veterinary Radiology application}} This paper introduces a novel framework for identifying aberrant predictions. It goes beyond traditional metrics and uncertainty quantification techniques to find classifications that could have significant adverse real-world consequences. Particularly, we apply this framework to the domain of veterinary radiology which remains underexplored within the machine learning community.
\item{\textbf{Enhancement of Model Value}} By identifying aberrant predictions, our framework allows a model to make robust, trustworthy predictions. This allows it to be deployed in real-world situations with reasonable failure-modes, which is a necessity in many application areas.

\end{enumerate}

\section{Related work}

Several approaches have been proposed to address the challenges posed by ambiguous or uncertain labels, or to make sure models are well-calibrated in their predictions. This section explores prior work in reliability, robustness, and calibration in the context of aberrant predictions.

\textbf{Classifier Calibration}:
The study of model calibration explores the reliability of model predictions, first explored by \cite{guo2017}. Calibration has been well-defined in recent years by \cite{wang2023}; In particular, a well-calibrated model predicts class probabilities that faithfully estimate the true correctness likelihood of their predictions. A model is perfectly calibrated if for some data distribution $\mathcal{D}$, for all input pairs $(x, y) \in \mathcal{D}$, if a model predicts $p_i = 0.8$, then 80\% such pairs have $y_i$ as a ground truth label. Work on calibration can be broadly categorized into post-hoc methods which calibrate models after training (\cite{mozafari2018}, \cite{minderer2021}, \cite{kull2019}), regularization methods during training (\cite{pereyra2017}, \cite{lin2017}, \cite{kumar2018}, \cite{bohdal2021}), data augmentation methods (\cite{muller2019}, \cite{thulasidasan2019}), and alleviating miscalibration by injecting randomness with uncertainty estimation (\cite{blundell2015}, \cite{wang2021}, \cite{pei2022}). However, even a perfectly-calibrated model can make aberrant predictions.

\textbf{Uncertainty quantification}:
Uncertainty quantification (UQ) aims to differentiate between various sources of uncertainty in model predictions \cite{moloud2020}. Uncertainty can be broadly modeled with Bayesian techniques (\cite{Brach2020SingleSM}, \cite{Wang2018AleatoricUE}, \cite{Liu2019UniversalAP}) and ensemble learning techniques (\cite{Lakshminarayanan2016SimpleAS}, ). \cite{malinin2018} makes a distinction between \textit{distributional uncertainty, data uncertainty}, and \textit{model uncertainty} to estimate the distributional uncertainty. \cite{kendall2017} defines \textit{aleatoric} uncertainties to capture noise inherent in observations, and \textit{epistemic} uncertainties as accounting for uncertainty in the model, which disappears with enough data. Although UQ provides a nuanced understanding of the types of uncertainties associated with model predictions, it does not address their nature: a model can make a low-uncertainty aberrant prediction.

\textbf{Conformal Predictions}:
Conformal predictions extend the idea of uncertainty quantification by constructing prediction regions that contain the ground truth with a specified degree of confidence \cite{Angelopoulos2021AGI}. Recent work in conformal prediction focuses on procedures with good performance on particular desiderata, like small set sizes \cite{Sadinle2016LeastAS}, balanced coverage across features space \cite{Angelopoulos2020UncertaintySF}, \cite{Romano2020ClassificationWV}, \cite{Cauchois2020KnowingWY} and errors balanced across classes \cite{Sadinle2016LeastAS} \cite{Guan2019PredictionAO} \cite{Lei2014ClassificationWC}. However, like UQ, conformal predictions only provide a measure of \textit{uncertainty about ground truth} and do not delve into situations where the model's output corresponds to the ground truth, but in which the information the model used to provide that prediction was aberrant.

These methodologies primarily focus on quantifying predictive uncertainty or ensuring a certain level of reliability in model predictions. However, none venture into analyzing what the kinds of predictions the models make say about what parts of the data the model is using, and thus in which situations predictions become unreliable. A model's performance is ultimately bounded by the diversity and size of the data it was trained on and the capacity of its architecture, and even well-calibrated and aligned models can make aberrant predictions. This is particularly problematic in complex, real-world scenarios where anomalies and rare events can occur.

The literature on this issue remains relatively sparse. The exploration of aberrant predictions, their identification, and their mitigation form the cornerstone of our investigation. A thorough understanding of these types of predictions enhances the practical utility and trustworthiness of machine learning models in real-world applications, particularly in domains laden with label uncertainty and which demand predictive trust.

\section{Methods}
We propose a strategy to improve an AI-aided veterinary radiology system by finding aberrant predictions. The system is designed to identify and suggest potential fracture regions in radiographic films, expediting diagnosis. Regions identified as suspicious, even if ultimately incorrect, hold value for clinicians. However, aberrant predictions risk eroding their trust in the technology. To address this, we leverage an existing production solution as our baseline, which uses a limb-fracture classifier built on the EfficientNet V2 XL architecture.\cite{Tan2021}.

\subsection{Workflow}
\begin{figure}[h]
    \centering
    \includegraphics[width=\linewidth]{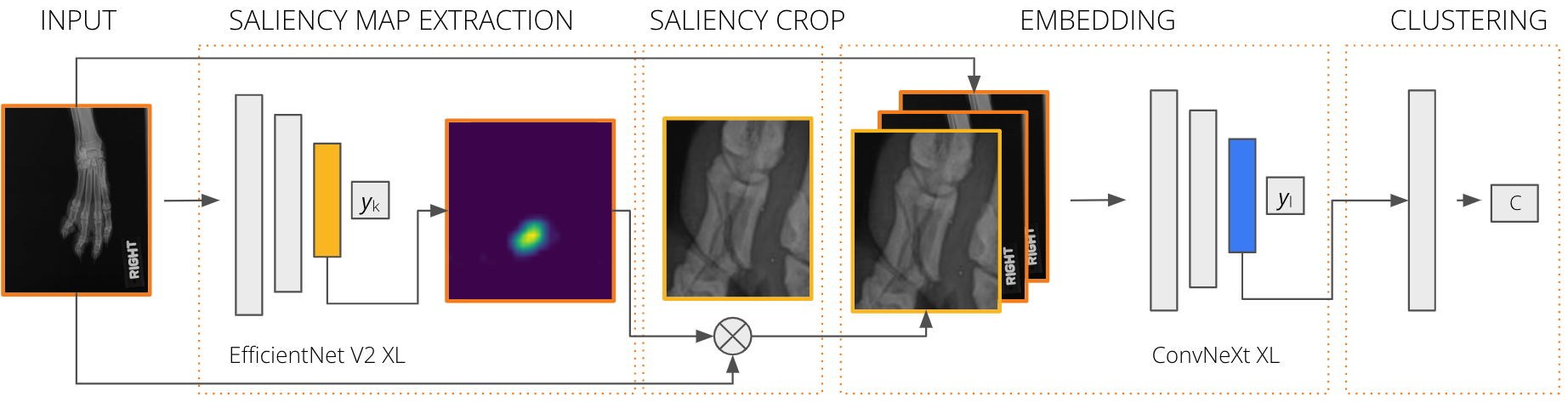}
    \caption{Workflow outlining the steps to identify aberrant predictions. $C$ represents the cluster. $y_k$ and $y_l$ are the classification results.}
    \label{fig:pipeline}
\end{figure}

We show our framework in Figure \ref{fig:pipeline}. We first generate saliency maps from the production classifier to crop the original radiographic images. These cropped regions are then embedded and subjected to unsupervised clustering to distinguish between logical and aberrant predictions. This methodology allows us to move beyond simply asking where the fracture is, and instead try to determine whether the highlighted region logically resembles a fractured area. 

\subsubsection*{Saliency maps}
Saliency maps are employed to discern the regions in the images that are pivotal in driving the classifier's decisions \cite{Adebayo2020}. We employed Grad-CAM \cite{Ramprasaath2019} for generating saliency maps due to its ease of application, despite certain cited shortcomings  \cite{Adebayo2020}. It is pertinent to note that while Grad-CAM sufficed for our objectives, alternative methods like SHAP (SHapley Additive exPlanations) could be explored in other projects for potentially more nuanced insights \cite{Lundberg2017, Teneggi2022}.

\subsubsection*{Input preprocessing}
The original images are cropped based on the generated saliency maps to isolate the discriminative regions influencing the classifier. The cropped images and saliency maps are then concatenated along the channel dimension to encapsulate the pertinent information for further analysis.

\subsubsection*{Aberrant prediction classifier}
An encoder model, built on a ConvNeXt architecture \cite{Liu2022}, is utilized to embed the concatenated images into a lower-dimensional space. Although our approach employs supervised learning with a small labeled dataset to fine-tune the embeddings for identifying aberrant predictions, a Variational Autoencoder \cite{Higgins2017} or other unsupervised frameworks like SPICE \cite{Niu2021SPICESP} or SCAN \cite{VanGansbeke2020LearningTC} could be employed for a fully unsupervised approach, albeit potentially at the cost of lower performance \cite{Zhang2022ImprovingVR}.

\subsubsection*{Nearest neighbors clustering}
Post-embedding, a clustering step employing Nearest Neighbors (KNN) is performed to segregate the images into distinct groups, each indicative of either logical or aberrant classification predictions \cite{Chen2017UnsupervisedMC, Li2017DiscriminativelyBI, Xie2015UnsupervisedDE}. This method allows for discerning whether highlighted regions logically resemble fractured areas.

We employ a dual methodology to determine the ideal number of clusters. First, we conduct a systematic sweep over a range of clusters, and use Silhouette Score \cite{Shahapure2020ClusterQA} and Adjusted Rand Index Score \cite{Hubert1985} to quantitatively evaluate the clustering outcomes. This data-driven approach serves as the first filter in determining the optimal number of clusters. Second, we prioritize interpretability by manually inspecting the images to identify visual features that could serve as natural cluster centers. We aim to ensure that the clusters represent distinct, interpretable categories that reflect the underlying structure of the data. Through this inspection, we identify a set of major visual features that we believe the model should capture. These features include vertical lines, horizontal lines, oblique lines, absence of lines, zoomed-out images, and the presence of medical devices among others.

\begin{figure}[htbp]
    \centering
    \begin{subfigure}{0.3\textwidth}
        \centering
        \includegraphics[width=\textwidth]{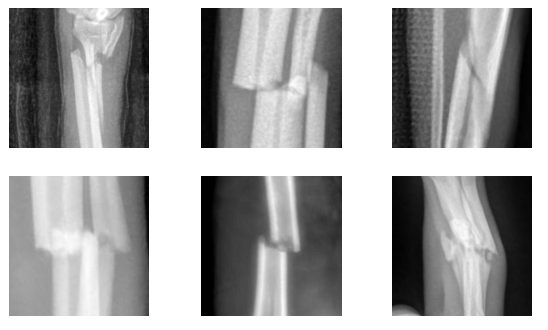}
        \caption{Vertical Lines}
    \end{subfigure}%
    \tikz{\draw[white, thick](0,0)--++(0.5cm,0);} 
    \begin{subfigure}{0.3\textwidth}
        \centering
        \includegraphics[width=\textwidth]{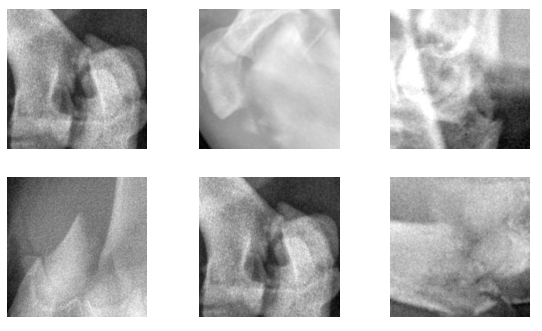}
        \caption{No Lines}
    \end{subfigure}%
    \tikz{\draw[white, thick](0,0)--++(0.5cm,0);} 
    \begin{subfigure}{0.3\textwidth}
        \centering
        \includegraphics[width=\textwidth]{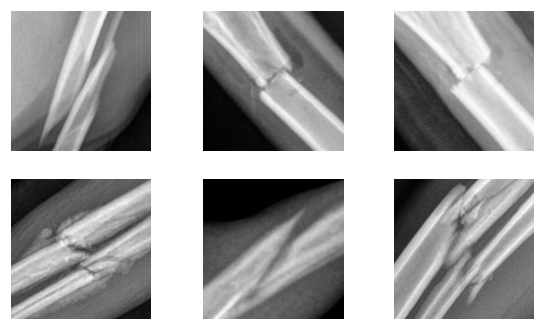}
        \caption{Oblique Lines}
    \end{subfigure}\\[1ex] 
    \begin{subfigure}{0.3\textwidth}
        \centering
        \includegraphics[width=\textwidth]{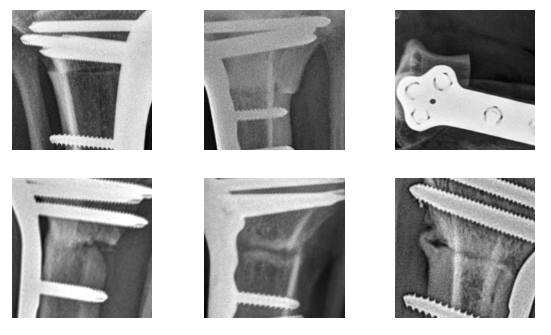}
        \caption{Medical Device}
    \end{subfigure}%
    \tikz{\draw[white, thick](0,0)--++(0.5cm,0);} 
    \begin{subfigure}{0.3\textwidth}
        \centering
        \includegraphics[width=\textwidth]{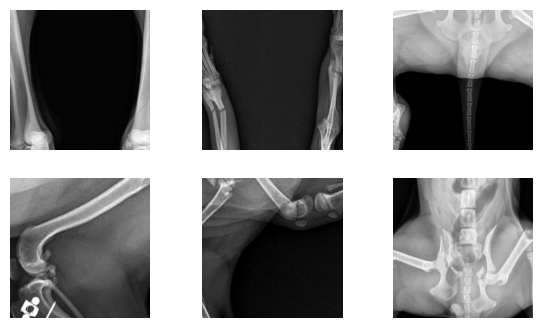}
        \caption{Zoomed Out}
    \end{subfigure}%
    \tikz{\draw[white, thick](0,0)--++(0.5cm,0);} 
    \begin{subfigure}{0.3\textwidth}
        \centering
        \includegraphics[width=\textwidth]{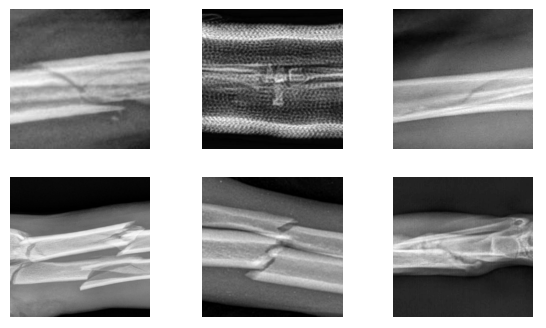}
        \caption{Horizontal Lines}
    \end{subfigure}
    \caption{Sampling of images belonging to each targeted cluster. Note that these images have not been used for training and are only aimed at providing a better understanding of the dataset and holistic cluster performance. }
    \label{fig:cluster_examples}
\end{figure}

We ultimately select six clusters. This choice aligns with the major visual features we wish the model to recognize, achieving a balance between optimization and interpretability. Figure \ref{fig:cluster_examples} showcases a representative sampling of images belonging to each cluster. This categorization helps in understanding the nature of the classifier's outputs and enriches insights into whether the clusters are being formed based on features that are both interpretable and logically coherent.

This workflow allows us to more deeply understand a classifier's behavior, especially in the realm of aberrant predictions, and paves the way for more interpretable and reliable machine learning models in medical imaging applications.

\subsection{Data}
The study utilizes a dataset comprising 3970 veterinary radiology images associated with limb fractures (80/20 train test split). The identification of aberrant predictions is facilitated through the analysis of regions within the images that significantly influence the classifier's predictions. To further elucidate the framework's capability in identifying aberrant predictions, two complementary evaluation methods were employed. 
\begin{enumerate}
    \item\textbf{Saliency-map-crop evaluation}
    The saliency maps were initially evaluated using fracture mask bounding box labels provided by expert veterinarians. The metric we employ, which we refer to as \textit{Saliency Crop Accuracy} (SCA) is a specialized adaptation of the Intersection over Union (IoU) metric commonly used for object detection tasks \cite{Rezatofighi2019GeneralizedIO}. Unlike traditional IoU focusing solely on overlapping areas between predicted and ground-truth bounding boxes, SCA considers the centroids' distance when no overlap occurs, introducing a decay factor proportional to the inverse of this distance. A high SCA value indicates a likely accurate prediction, while a zero value signifies misdirection in the saliency map. This evaluation suggests that all aberrant predictions will exhibit a very low saliency map accuracy metric value, which serves as a quantifiable measure of the aberrance of the predictions.
    
    \item\textbf{Visual inspection}
    However, the saliency map metric alone may not capture the complete picture, especially in scenarios where reasonable-looking regions did not contain a fracture (i.e. the SCA will be low, but the prediction is not aberrant). To address this limitation, a visual inspection of the predictions in each scenario was carried out. This dual-method evaluation motivates a more holistic approach to understanding `accuracy', emphasizing the importance of not only examining the alignment between the saliency maps and ground truth but also scrutinizing the logical coherence of the identified regions.  
\end{enumerate}

\section{Experiments}
The primary objective of the experiments is to validate the efficacy of the proposed framework in identifying aberrant predictions, leveraging expert evaluations as a benchmark. In addition, we add an interpretability analysis, based on the visual features handcrafted in Figure \ref{fig:cluster_examples}. Finally, we evaluate the production impact of the proposed solution.

\subsubsection*{Evaluate clusters based on Saliency Crop Accuracy (SCA)}
\begin{figure}[h]
    \centering
    \begin{minipage}{5.5cm}
        \textbf{(A)}
        \centering \\
        \includegraphics[width=\linewidth]{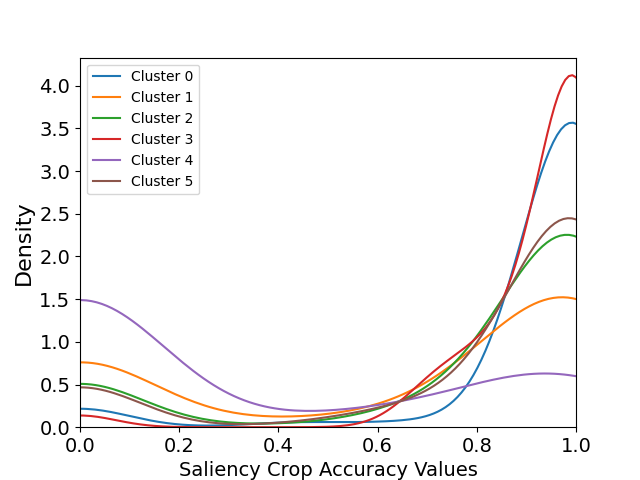}
        \label{fig:kde_a}  
    \end{minipage}
    \begin{minipage}{8cm}
        \textbf{(B)} 
        \centering \\
        \includegraphics[width=\linewidth]{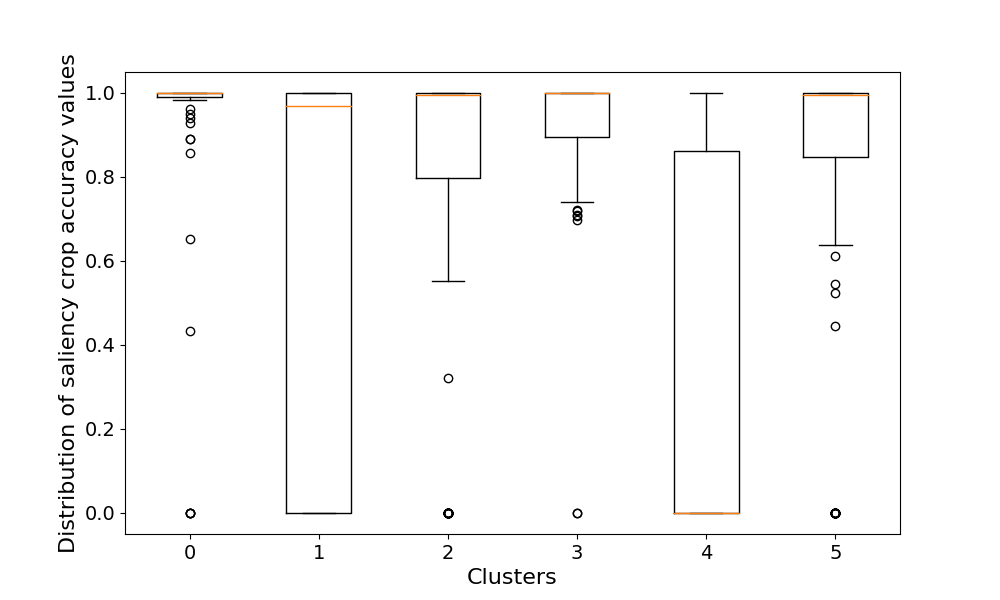}
        \label{fig:boxplot_b}  
    \end{minipage}
    \caption{(\textbf{A}) KDE of clusters with respect to expert evaluations. Distinct peaks or modes in the density distribution indicate a high concentration of similar expert evaluations, potentially signaling the consistency of either logical or aberrant predictions within that cluster. This is the case for cluster 4, for instance, which shows a high concentration of aberrant predictions (low SCA), or cluster 0, showing the opposite. (\textbf{B}) Box-plot showing the distribution of salience map accuracy for each cluster. The box-plot echoes the insights from the KDE plot. Specifically, clusters 1 and 4 stand out for containing a significant proportion of aberrant predictions, reinforcing the observations made through the density distribution analysis.}
    \label{fig:kde}
\end{figure}

To assess the framework's capability to identify aberrant predictions, we examined the density distribution of each cluster concerning the expert evaluations leveraging Kernel Density Estimations (KDE) \cite{Modak2022ANM} (Figure \ref{fig:kde}A). The underlying hypothesis is that clusters formed around images where the saliency map contains aberrant predictions should show a distinct density distribution when contrasted with clusters formed around images where the saliency map accurately indicates a fractured region. 

A box-plot depicting the distribution of accuracy for each cluster was generated (Figure \ref{fig:kde}B). It was observed that clusters containing aberrant predictions displayed lower accuracies, suggesting the presence of aberrant predictions, while clusters devoid of them had higher accuracies, indicating logical predictions. This analysis substantiated the capability of our framework in segregating aberrant predictions from logical predictions, aligning with expert evaluations.

\subsubsection*{Qualitative human evaluation}
\begin{figure}[htbp]
    \centering
    \begin{subfigure}{0.3\textwidth}
        \centering
        \includegraphics[width=\textwidth]{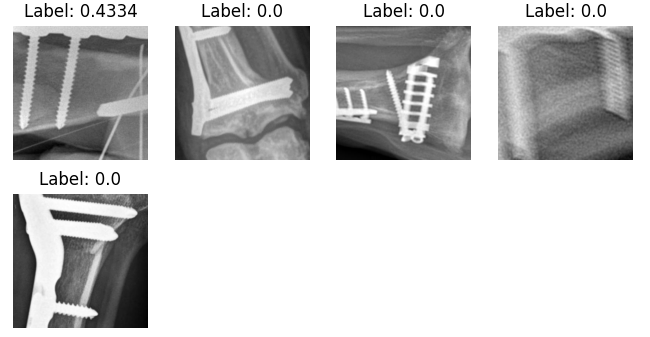}
        \caption{Cluster 0}
    \end{subfigure}%
    \tikz{\draw[white, thick](0,0)--++(0.1cm,0);} 
    \begin{subfigure}{0.3\textwidth}
        \centering
        \includegraphics[width=\textwidth]{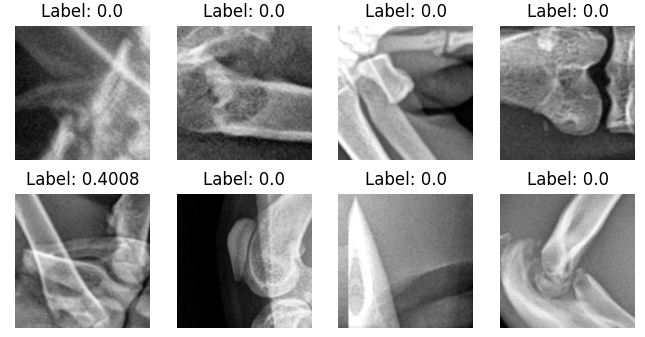}
        \caption{Cluster 1}
    \end{subfigure}%
    \tikz{\draw[white, thick](0,0)--++(0.1cm,0);} 
    \begin{subfigure}{0.3\textwidth}
        \centering
        \includegraphics[width=\textwidth]{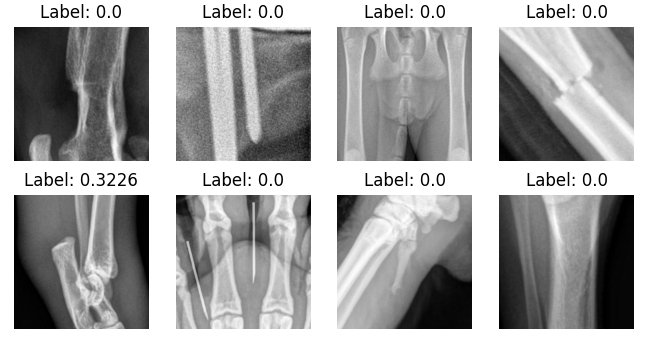}
        \caption{Cluster 2}
    \end{subfigure}\\[1ex] 
    \begin{subfigure}{0.3\textwidth}
        \centering
        \includegraphics[width=\textwidth]{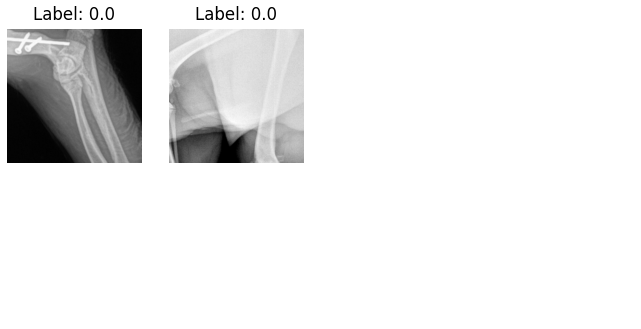}
        \caption{Cluster 3}
    \end{subfigure}
    \tikz{\draw[white, thick](0,0)--++(0.1cm,0);} 
    \begin{subfigure}{0.3\textwidth}
        \centering
        \includegraphics[width=\textwidth]{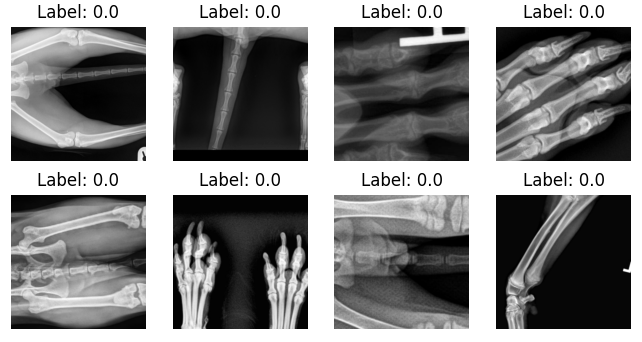}
        \caption{Cluster 4}
    \end{subfigure}%
    \tikz{\draw[white, thick](0,0)--++(0.1cm,0);} 
    \begin{subfigure}{0.3\textwidth}
        \centering
        \includegraphics[width=\textwidth]{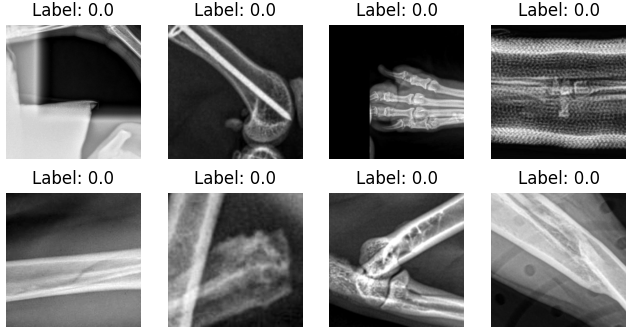}
        \caption{Cluster 5}
    \end{subfigure}%
    \caption{Examples of images with low-SCA values for each cluster. Clusters 0 and 3 have fewer than 8 mistake examples. These images serve as illustrative samples of what low-SCA images from each cluster would look like}
    \label{fig:cluster_mistakes}
\end{figure}

A qualitative evaluation was conducted to ascertain the practical relevance and justifiability of the mistakes made by the original production classifier. A sampling of the low SCA-value images from each cluster was reviewed by expert radiologists based on whether the highlighted regions were justifiably located for fracture assessment. Figure \ref{fig:cluster_mistakes} shows an example of low-SCA crops for each cluster to illustrate this process. Note that not all clusters contain over 8 low-SCA crops for the display.

The expert radiologists provided their assessments on the nature of the mistakes and the relevance of the identified fractures. For instance, it was unanimously agreed that all the fractures identified in cluster 0 were of interest and held significant value for the business production scenario, particularly with respect to aiding accurate and trusworthy fracture diagnosis. The aberrant predictions rate was below 1\% for cluster 3 and below 5\% for clusters 2 and 5. Clusters 4 and 1 contain most of the aberrant predictions with 81\% for cluster 4 and 25\% for cluster 1.

\subsubsection*{Increased interpretability of the results}
\begin{figure}[htbp]
    \centering
    \begin{subfigure}{0.3\textwidth}
        \centering
        \includegraphics[width=\textwidth]{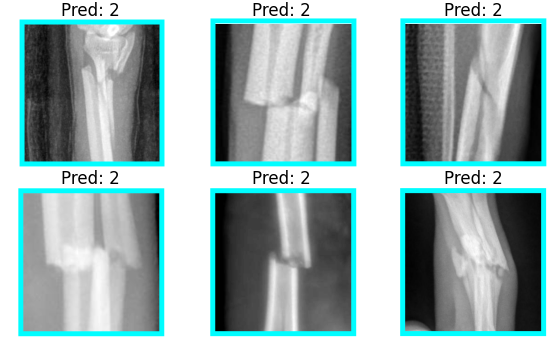}
        \caption{Vertical Lines}
    \end{subfigure}%
    \tikz{\draw[white, thick](0,0)--++(0.5cm,0);} 
    \begin{subfigure}{0.3\textwidth}
        \centering
        \includegraphics[width=\textwidth]{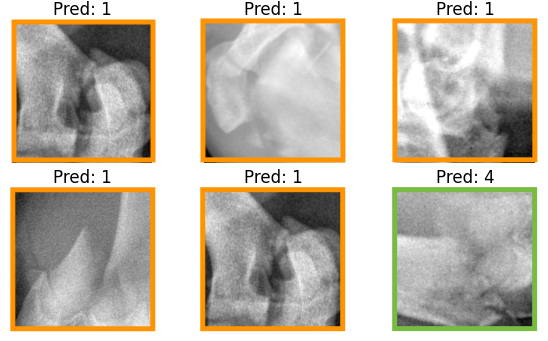}
        \caption{No Lines}
    \end{subfigure}%
    \tikz{\draw[white, thick](0,0)--++(0.5cm,0);} 
    \begin{subfigure}{0.3\textwidth}
        \centering
        \includegraphics[width=\textwidth]{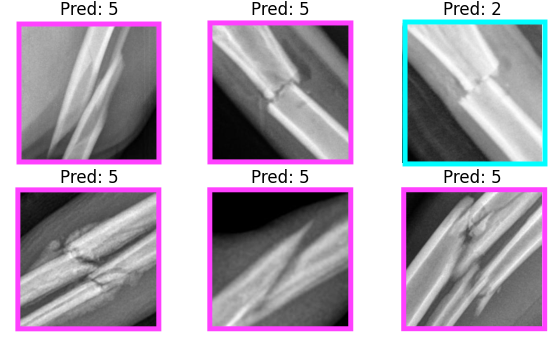}
        \caption{Oblique Lines}
    \end{subfigure}\\[1ex] 
    \begin{subfigure}{0.3\textwidth}
        \centering
        \includegraphics[width=\textwidth]{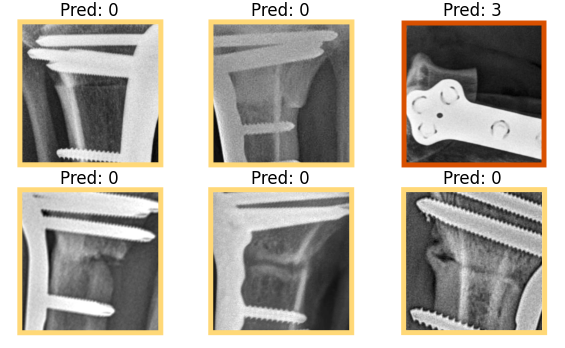}
        \caption{Medical Device}
    \end{subfigure}%
    \tikz{\draw[white, thick](0,0)--++(0.5cm,0);} 
    \begin{subfigure}{0.3\textwidth}
        \centering
        \includegraphics[width=\textwidth]{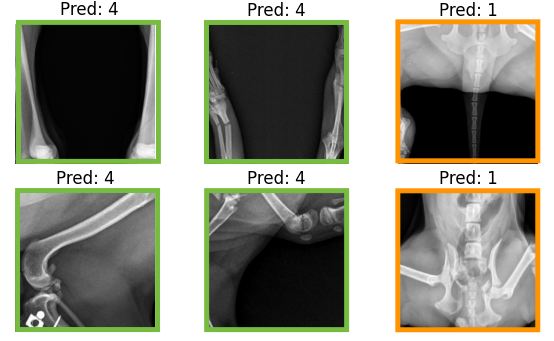}
        \caption{Zoomed Out}
    \end{subfigure}
    \tikz{\draw[white, thick](0,0)--++(0.5cm,0);} 
    \begin{subfigure}{0.3\textwidth}
        \centering
        \includegraphics[width=\textwidth]{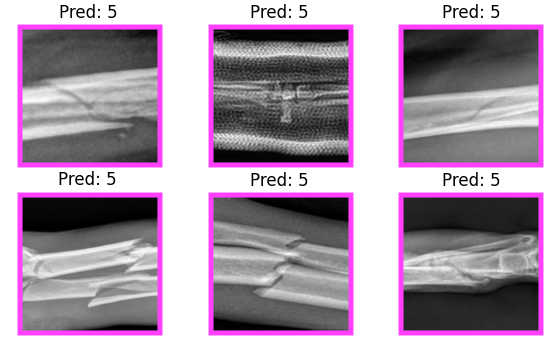}
        \caption{Horizontal Lines}
    \end{subfigure}
    \caption{Performance of the clustering system on saliency crops with different visual features. The title of each sub-image indicates its final predicted cluster. Different colors are used to highlight the diverse clusters. This visualization aids in understanding how the images were ultimately categorized by the model.}
    \label{fig:final_preds_clusters}
\end{figure}

To obtain a more holistic understanding of model performance, we use 6 major visual features identified during data analysis as a guide for understanding the model's decision-making process. Figure \ref{fig:final_preds_clusters} displays the performance of the clustering system on example-saliency crops with each one of the major visual features. 

All of the saliency crops selected as representative of vertical lines were assigned by the KNN to cluster 2, and all the horizontal lines to cluster 5, along with most oblique lines. Interestingly, different views of medical devices appear to be clustered separately, with frontal views of the plate found in cluster 3 and crops containing the screws in cluster 0. The zoomed out crops, which mostly represent aberrant mistakes, are found in cluster 4. Finally, the crops without lines are mostly found in cluster 1.

\subsubsection*{Production Evaluation}
 While precision and recall fail to completely describe the performance of the model, they can serve as valuable benchmarks for getting a rough intuition in model evaluations. Especially when accounting for the impact of aberrant mistakes in them.
 
\begin{figure}[h]
    \centering
    \includegraphics[width=0.6\linewidth]{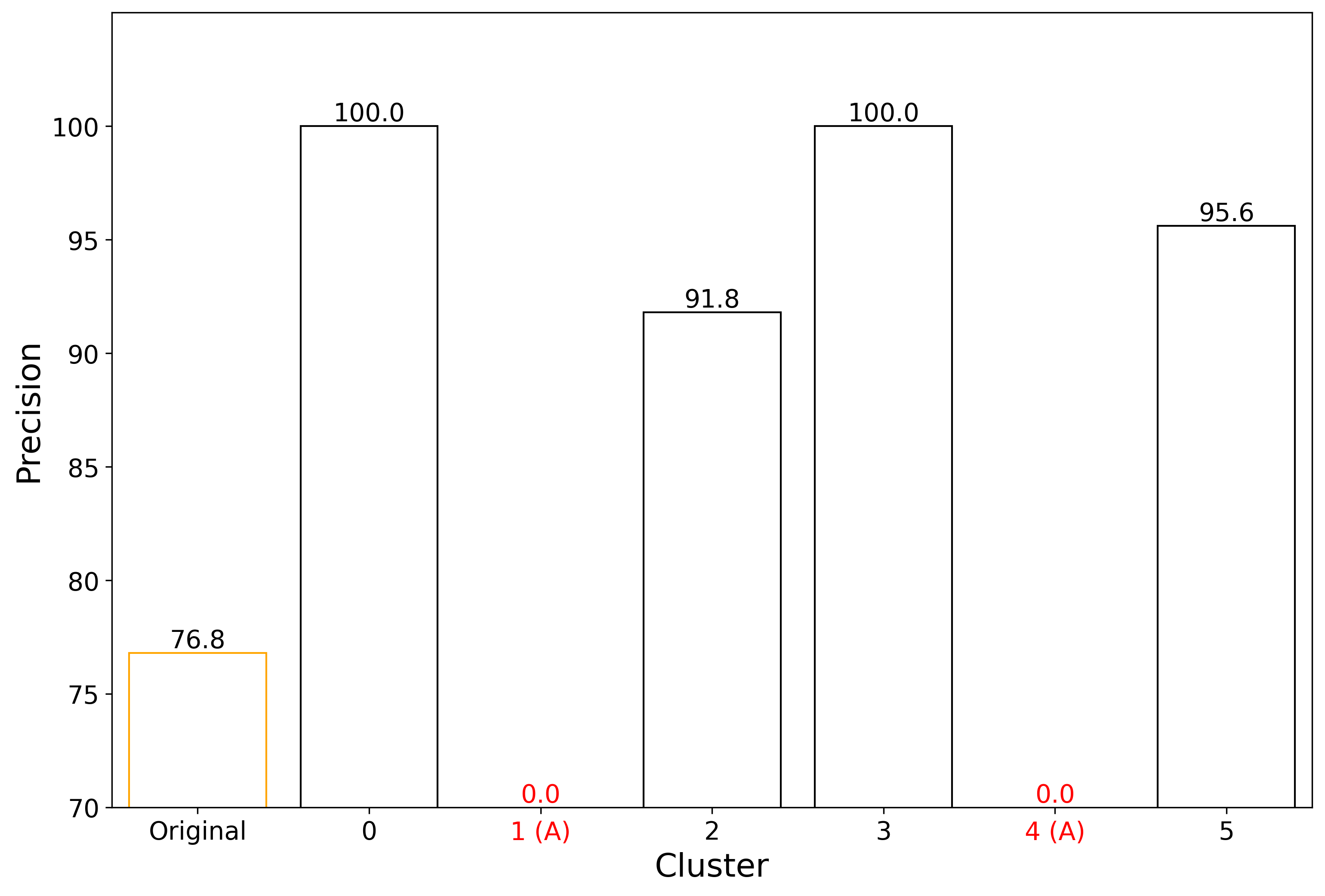}
    \caption{Cluster precision after evaluation in the production model. The precision in clusters 0, 2, 3, and 5 is dramatically better than the original model. Clusters 1 and 4 contained mostly aberrant predictions and have a precision of 0\%, with no true positives.}
    \label{fig:precision}
\end{figure}

 Under this light, we compare the new method with simply using the previously existing classifier approach (used to generate the saliency maps). We show our results in Figure \ref{fig:precision}. The original model had a precision of 76.8\%. The final precision results for the best model pushed to production are 91.5\% for cluster 1, 91.8\% for cluster 2, 100\% for clusters 0 and 3, and 95.6\% for cluster 5. Compared to the previously existing model this implies an improvement of up to over 20\% for all the clusters without aberrant predictions. In turn, recall, excluding aberrant predictions, is only reduced by 7.2\%. 

The increase in overall precision as a result of this method is good but not the main thing we want to emphasize. The primary value-add is that the types of false positives are better; the mistakes become truly ambiguous, rather than blatantly wrong.

\section{Discussion}

The experiments in Figure \ref{fig:kde} reveal that using KDE and box plots in tandem is effective for understanding the behavior of our model. The distinct distribution patterns observed in different clusters support the idea of excluding clusters with a high concentration of aberrant predictions (clusters 1 and 4) from the production pipeline. 

The proposed framework significantly improved precision by 20\%, as shown in the Production Evaluation. This increase in precision in clusters `devoid' of aberrant predictions illustrates the effectiveness of our methodology. The comparison with the previous model, which had a precision of 76.8\%, demonstrates the robustness and applicability of our framework in a production environment. The qualitative human evaluation further validated the effectiveness of our framework. The agreement among expert radiologists, especially on images from cluster 0, highlights the practical utility and potential real-world impact of our framework in veterinary radiology.

In summary, the experiments validate our framework's capability in not only identifying fracture locations but also in ensuring the logical accuracy of the highlighted regions, addressing a critical aspect in veterinary radiology machine learning applications.

\subsection{Limitations}
A significant drawback of our method is the need for manual re-clustering after model retraining. This manual intervention not only introduces a resource and time overhead but may also introduce human bias into the process. We also rely on the assumption that clusters with a high concentration of aberrant predictions should be excluded from the production pipeline. There may be situations in which doing this has lower effects on standard metrics like precision and recall. There are ways to mitigate this -- for instance, by clustering based on the performance of a test set -- but this is at the cost of losing interpretability, since we won't know what's in the new clusters.

Lastly, the generalizability of the framework across different domains of machine learning remains to be fully explored. While we advocate for the broad applicability of aberrance-based methods, the efficacy of the proposed framework in other domains and the manual intervention required for re-clustering post-model retraining may pose challenges to its widespread adoption.

\section{Conclusion}

The paper introduces and defines the concept of aberrant predictions, offering a new lens with which to evaluate machine learning models beyond traditional metrics. We develop a novel framework which uses the concept to significantly improve model precision and interpretability in veterinary radiology, fostering a better understanding between ML models and medical practitioners. This methodology lays a strong foundation for future research aimed at minimizing aberrant predictions and enhancing the real-world clinical utility of ML models. We use veterinary radiology as a case study to showcase the utility of our method, but we argue that aberrance-based methods have high practical utility across many domains of machine learning.

\begin{ack}

This paper is supported by SignalPET, which owns all data and resources used for this research. Special acknowledgment goes to Lior Uzan and Bijan Varjavand, whose insightful discussions were instrumental in shaping this article.
\end{ack}

\section*{References}
\printbibliography[heading=none]

\end{document}